# FWB-NET: FRONT WHITE BALANCE NETWORK FOR COLOR SHIFT CORRECTION IN SINGLE IMAGE DEHAZING VIA ATMOSPHERIC LIGHT ESTIMATION

*Cong Wang[1], Yan Huang[1], Yuexian Zou[1,2*], Yong Xu[3]*

[1] ADSPLAB, School of ECE, Peking University, Shenzhen, China
[2] Peng Cheng Laboratory, Shenzhen, China
[3] School of Computer Science and Engineering, South China University of Technology, Guangzhou, China
*Corresponding author: zouyx@pku.edu.cn

## ABSTRACT

In recent years, single image dehazing deep models based on Atmospheric Scattering Model (ASM) have achieved remarkable results. But the dehazing outputs of those models suffer from color shift. Analyzing the ASM model shows that the atmospheric light factor (ALF) is set as a scalar which indicates ALF is constant for whole image. However, for images taken in real-world, the illumination is not uniformly distributed over whole image which brings model mismatch and possibly results in color shift of the deep models using ASM. Bearing this in mind, in this study, first, a new non-homogeneous atmospheric scattering model (NH-ASM) is proposed for improving image modeling of hazy images taken under complex illumination conditions. Second, a new U-Net based front white balance module (FWB-Module) is dedicatedly designed to correct color shift before generating dehazing result via atmospheric light estimation. Third, a new FWB loss is innovatively developed for training FWB-Module, which imposes penalty on color shift. In the end, based on NH-ASM and front white balance technology, an end-to-end CNN-based color-shift-restraining dehazing network is developed, termed as FWB-Net. Experimental results demonstrate the effectiveness and superiority of our proposed FWB-Net for dehazing on both synthetic and real-world images.

*Index Terms—* Single image dehazing, atmospheric scattering model, color shift correction, front white balance

## 1. INTRODUCTION

Single image dehazing aims to recover the clean image from a hazy input, which has received significant attention in the vision community over the past few years. According to the atmospheric scattering model (ASM) [1, 2, 3], the hazing process is usually formulated as:

$$I(x,y) = J(x,y)t(x,y) + A(1 - t(x,y)) \quad (1)$$
$$t(x,y) = e^{-\beta d(x,y)} \quad (2)$$

where I(x, y) is the hazy image. J(x, y) is the haze-free image. $\beta$ is a coefficient which represent the degree of haze. d(x, y) denotes the distance from object to camera. t(x, y) is transmission map. A is global atmospheric light factor, which can be considered as a constant among the whole image. Given a hazy image, most dehazing algorithms try to estimate t and A, and calculate clean image J by equation (3):

$$J(x,y) = \frac{I(x,y) - A}{t(x,y)} + A \quad (3)$$

However, it has been observed that the atmospheric light factor A is not globally constant under any conditions, such as there are both shadow and bright areas in one image. This deviation in modeling causes some bias in dehazing results. To modeling more general dehazing scenes with non-homogeneous illumination, we modify the ASM and correcting the atmospheric light factor to be related to pixel position on image. This model is termed as non-homogeneous atmospheric scattering model (NH-ASM).

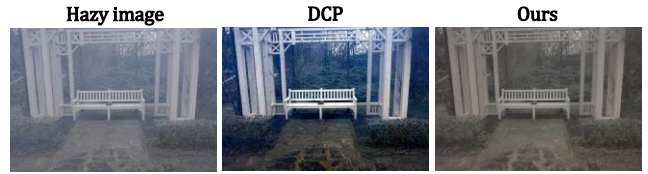

**Fig. 1.** The dehazing results with different methods. Our method can get clean results without color shift.

As for single image dehazing methods, they can be categorized as: image prior-based methods and deep learning-based methods. Image prior-based methods try to find prior information that can be definitive in improving visibility of hazy images. Such as DCP [4] and Berman [5] estimate the transmission map t based on dark channel prior and haze-line prior, then getting haze-free image by equation (3). Deep learning-based methods try to estimate parameters of ASM by neural networks. DehazeNet [6] use CNN [7] to obtain multi-scale features and output transmission. AOD-Net [8], GFN [9], GDN [10] and ABC-Net [11] are end-to-end neural networks which can output haze-free image directly. In particular, ABC-Net can effectively restrain local color shift of dehazing result via specially designed loss and activation function.

However, all these dehazing algorithms focus on solving the transmission, but using very rough methods to solve the atmospheric light factor. It causes color shift issue in dehazing result due to inaccurate atmospheric light estimation. This kind of color shift diffuses to the whole image and expresses especially serious in real-world dehazing scene. This phenomenon is similar to the visual characteristic of images taken in a colored light environment. The technique, which restores the original color of objects from an image taken in a colored light environment, is called white balance [12]. The core of white balance is to accurately estimate the color of light from an image, and using it to adjust the color of objects in the image. As for dehazing, the atmospheric light factor in ASM reflects the comprehensive influence of light. Therefore, in order to restrain the color shift of dehazing result, it only needs to estimate the color of light by accurately calculating the atmospheric light factor.

Based on the above motivation, we propose a pixel level, U-Net-based [13] atmospheric light estimation module, and design an exclusively color-shift-restraining loss function for training this module. This module can accurately estimate the atmospheric light factor of an input hazy image pixel by pixel, and restrain the diffuse color shift of dehazing result. It plays a similar role in picture editing. And this module has already completed the adjustment for color before creating dehazing results, instead the traditional white balance technology adjusts the color after generating images. So, we term it as front white balance module.

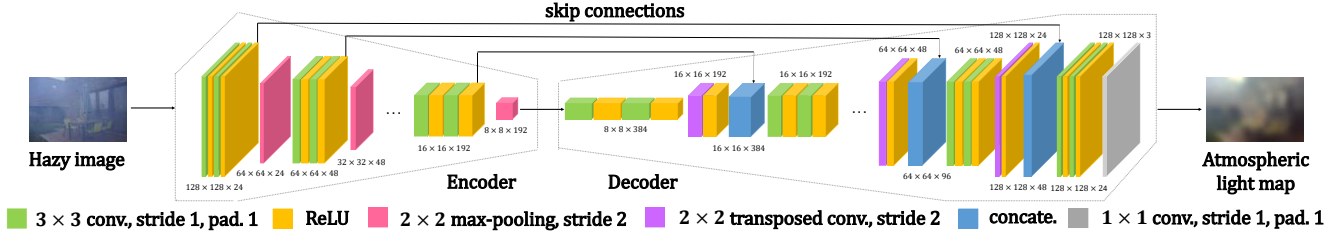

**Fig. 2.** The architecture of FWB-Module.

Based on the non-homogeneous atmospheric scattering model (NH-ASM), we propose an end-to-end single image dehazing network with restraining color shift, which applies front white balance technology. Experiments show that our method is capable of keeping original colors of objects in dehazing images. The major contributions of our work are:

(1) We propose a non-homogeneous atmospheric scattering model (NH-ASM), which models atmospheric light factor as related to pixel index in image. It can express the imaging law of real-world hazy scene better.

(2) We propose a new pixel-level atmospheric light estimation module, which can effectively limit the global color shift of dehazing. It is termed as front white balance module (FWB-Module), as it plays a similar role to white balance before generating the dehazing result.

(3) We propose a new loss function for training FWB-Module, termed as FWB-Loss. It can impose penalty on color shift to keep the original color of objects in dehazing result.

(4) We propose an end-to-end CNN-based dehazing network, termed as FWB-Net, which applies FWB-Module and FWB-Loss. The experimental results on both synthetic and real-world hazy databases provide strong support for the effectiveness of our proposed method.

## 2. PROPOSED METHOD

### 2.1. Non-homogeneous atmospheric scattering model

The classic form of atmospheric scattering model (ASM) is showed in equations (1)(2). In ASM, the atmospheric light factor A is a global constant, which is in line with the illumination characteristics of open field under sunlight. However, when there are shadows or multiple light sources in one scene, the optical property of different pixels in the image are obviously distinct. Therefore, we modify the ASM by changing the atmospheric light factor as position dependent. The new model is termed as non-homogeneous atmospheric scattering model (NH-ASM), which can describe the diverse and complex illumination conditions in real world more precisely. And the form of NH-ASM is showed in equations (4):

$$I(x,y) = J(x,y)t(x,y) + A(x,y)(1 - t(x,y)) \quad (4)$$

### 2.2. Front white balance module (FWB-Module)

According to NH-ASM (equation 4), there is a pixel-to-pixel and linear mapping relationship between hazy image I and atmospheric light map A. And it is suitable for modeling this mapping by an encoder-decoder model. Specifically, we use a U-Net [13] architecture with multi-scale skip connections between the encoder and decoder. Our framework is shown on Fig. 2, and it consists of two main units: the first is a 4-level encoder unit that is responsible for extracting a multi-scale latent representation of our input image; the second is a 4-level decoder unit, which has a bottleneck and transposed convolutional layers. At the first level of our encoder and decoder, the conv layers have 24 channels. For each subsequent level, the number of channels is doubled (i.e., the fourth level has 192 channels for each conv layer).

This module can learn the mapping between hazy image and atmospheric light factor, and getting pixel level atmospheric light maps. Due to the diffuse color shift in an image can be considered as the illumination of imaging has chromatic color. The atmospheric light estimate module can restrain the global distortion caused by illumination's color of dehazing result, so it has the similar effect as white balance. Moreover, this module has already corrected color shift before generating dehazing results, instead of traditional white balance technology, which adjusts the color after creating images. So, we term it as front white balance module (FWB-Module).

### 2.3. Front white balance loss function (FWB-Loss)

Color space is an abstract mathematical model, which describes the specific organization of colors. In color space, colors can be represented as tuples of numbers. Except the well-known RGB color space, there are CMYK, CIELAB [14] as common color spaces. Unlike RGB and CMYK, CIELAB is designed to approximate human vision. It aspires to perceptual uniformity of color. Therefore, we convert the image data from RGB to CIELAB color space, and taking the minimum deviation between dehazing results and ground truth haze-free images in CIELAB as the optimization goal. It can restrain the color shift of dehazing results in vision better.

Denote the value of each channel in RGB color space at pixel position (x, y) on the dehazing image and ground truth haze-free image as $J_{est}(x,y,c)$ and $J_{gt}(x,y,c)$, where $c \in \{R, G, B\}$. In order to complete the conversion from RGB to CIELAB color space, it should transform the value of RGB into an auxiliary XYZ color space first, and then conversing the value from XYZ to CIELAB. This process can be expressed as equation (5, 6, 7) [15]:

$$\begin{bmatrix} J_{type}(x,y,X) \\ J_{type}(x,y,Y) \\ J_{type}(x,y,Z) \end{bmatrix} = \begin{bmatrix} 0.412 & 0.358 & 0.180 \\ 0.213 & 0.715 & 0.072 \\ 0.019 & 0.119 & 0.950 \end{bmatrix} \begin{bmatrix} J_{type}(x,y,R) \\ J_{type}(x,y,G) \\ J_{type}(x,y,B) \end{bmatrix} \quad (5)$$

$$\begin{bmatrix} J_{type}(x,y,L) \\ J_{type}(x,y,a) \\ J_{type}(x,y,b) \end{bmatrix} = \begin{bmatrix} 0 & 116 & 0 \\ 500 & -500 & 0 \\ 0 & 200 & -200 \end{bmatrix} \begin{bmatrix} f[J_{type}(x,y,X)] \\ f[J_{type}(x,y,Y)] \\ f[J_{type}(x,y,Z)] \end{bmatrix} - \begin{bmatrix} 16 \\ 0 \\ 0 \end{bmatrix} \quad (6)$$

$$f(t) = \begin{cases} t^{\frac{1}{3}} & , t > \left(\frac{6}{29}\right)^3 \\ \frac{1}{3}\left(\frac{29}{6}\right)^2 \cdot t + \frac{4}{29} & , otherwise \end{cases} \quad (7)$$

where type $\in \{est, gt\}$. And as our experiments show that 99.9999% of the color in CIELAB color space are greater than $\left(\frac{6}{29}\right)^3$, the equation (6) and (7) can be simplified as follow:

$$\begin{bmatrix} J_{type}(x,y,L) \\ J_{type}(x,y,a) \\ J_{type}(x,y,b) \end{bmatrix} = \begin{bmatrix} 0 & 116 & 0 \\ 500 & -500 & 0 \\ 0 & 200 & -200 \end{bmatrix} \begin{bmatrix} J_{type}(x,y,X)^{1/3} \\ J_{type}(x,y,Y)^{1/3} \\ J_{type}(x,y,Z)^{1/3} \end{bmatrix} - \begin{bmatrix} 16 \\ 0 \\ 0 \end{bmatrix} \quad (8)$$

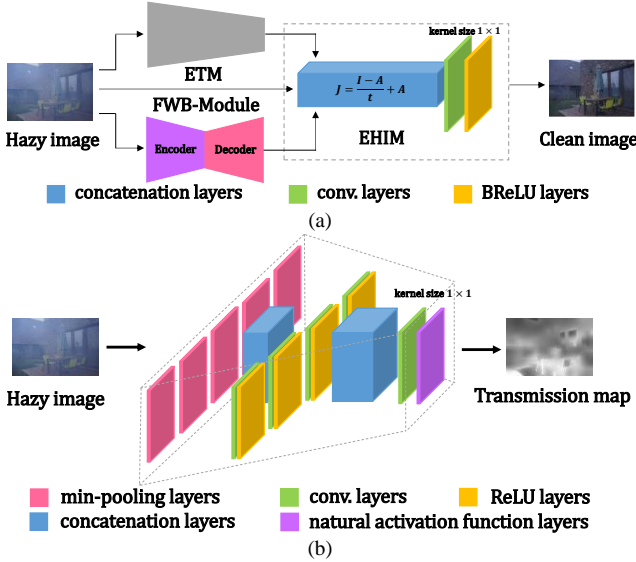

**Fig. 3.** (a) The network architecture of FWB-Net. (b) The architecture of ETM. The kernel sizes of min-pooling layers in the first line are $3 \times 3$, $5 \times 5$, $7 \times 7$, $9 \times 9$ and $11 \times 11$, respectively. And the kernel sizes of convolution layers in the second line are $3\times3$, $5\times5$, $7\times7$ (with dilation 2) and $7\times7$ (with dilation 3), respectively.

As the optimization goal is to minimize the distance between the estimated dehazing image $J_{est}$ and ground truth haze-free image $J_{gt}$ in CIELAB color space, so the loss function can be defined as:

$$\text{Loss}_J = \frac{1}{3MN} \sum_{x=1}^{M} \sum_{y=1}^{N} \sum_{k} [J_{est}(x,y,k) - J_{gt}(x,y,k)]^2 \quad (9)$$

where $k \in \{L, a, b\}$ means the data channel of CIELAB color space. From NH-ASM (equation 4), we have:

$$J(x,y,k) = \frac{t(x,y)-1}{t(x,y)} \cdot A(x,y,k) + \frac{I(x,y)}{t(x,y)} \quad (10)$$

Substituting equation (10) into equation (9), we can get the following equation:

$$\text{Loss}_A = \frac{1}{3MN} \sum_{x=1}^{M} \sum_{y=1}^{N} \left(1 - \frac{1}{t(x,y)}\right)^2 \sum_{k} [A_{est}(x,y,k) - A_{gt}(x,y,k)]^2 \quad (11)$$

Since $(1 - 1/t)^2$ is irrelevant to atmospheric light factor A as well as always positive, we omit it and get a simplified loss function:

$$\text{Loss}(A_{est}, A_{gt}) = \frac{1}{3MN} \sum_{x=1}^{M} \sum_{y=1}^{N} \sum_{k} [A_{est}(x,y,k) - A_{gt}(x,y,k)]^2 \quad (12)$$

Using equation (5) and (8) to convert the values of 3 channels of atmospheric light from RGB into CIELAB color space, then penalize the deviation between estimated and ground truth atmospheric light in CIELAB by equation (12). This operation can correct the color shift during dehazing. Since the loss function is used to optimize the front white balance module, we term it as front white balance loss function (FWB-Loss).

### 2.4. Front white balance network (FWB-Net)

Following ABC-Net [11], we design an end-to-end dehazing network based on NH-ASM. For restraining color shift, we apply FWB-Module and FWB-Loss to our dehazing network. That is the reason why the whole network is termed as FWB-Net. The architecture of FWB-Net is shown on Fig. 3 (a). This network has 3 modules: 1) front white balance module (FWB-Module); 2) estimating transmission module (ETM); 3) estimating haze-free image module (EHIM). The detail of FWB-Net is described in below.

**Front white balance module (FWB-Module):** Its details are described in section 2.2. This module estimates the color of illumination by calculating atmospheric light factor, and realizes the pre-correction of dehazing results' color shift by this way.

**Estimate transmission module (ETM):** It reuses the ETM of ABC-Net. The module directly learns the mapping between haze image and transmission map through a multi-scale CNN network. In particular, the module also has excellent ability to restrain the color shift caused by inaccurate transmission estimation through specially designed loss function and activation function. It also helps the whole network to achieve color shift correction. The detail of ETM shows in Fig. 3 (b).

**Estimate haze-free image module (EHIM):** After obtaining transmission $t$ and atmosphere light factor $A$, the equation (3) is employed to figure out the haze-free image $J$. In order to refine the output and remove noise, we process $J$ with a convolutional layer followed by a BReLU activation function. The detail of EHIM shows in Fig. 3 (a).

Our proposed FWB-Net is delicately designed for single image dehazing with restraining color shift.

## 3. EVALUATIONS

### 3.1. Datasets and training details

Adequate training data is essential for CNN-based methods, but since it is very difficult to collect both clean and hazy images of a same scene, real world dehazing databases are very scarce. Lots of methods, such as AOD-Net [8], use ASM [1] to synthesize dehazing dataset. By convention, we use NYU-Depth-v2 [16] to synthesize our dataset. NYU-Depth-v2 provide 1449 clean images and theirs corresponding depth information. Referring to NH-ASM and setting β=0.35, we use 1000 images and the rest 449 images to create training and testing set, respectively.

At the same time, we randomly set the basic atmospheric light factor between 0.3 and 1.5 of each image, and apply a random disturbance of less than 20% to the atmospheric light factor of each pixel. Using this method, we synthesize the non-homogeneous atmospheric light map of each synthetic hazy image.

In training session, we use FWB-Loss and MSE loss to train the Front White Balance Module (FWB-Module) and estimating transmission module (ETM) under the supervision of ground truth atmospheric light maps and transmission maps, respectively. Then, we use MSE loss to train the whole FWB-Net, under the supervision of ground truth haze-free images. With the help of SGD algorithm [17], we take 100 iterations to train FWB-Module and ETM in parallel, and take another 100 iterations to train the whole network, then repeat the cycle until all the FWB-Module, ETM and FWB-Net converge. While in testing session, hazy images are input into the model and clean images are used for evaluation.

### 3.2. Evaluations on synthetic hazy image dataset

We evaluate ABC-Net on a synthetic hazy dataset, which is obtained by setting β=0.35 and A in the range of [0.3, 1.5], randomly. Another six high-performance methods [4-6, 8-10] are reimplemented for comparing. The results are shown in Fig. 5(a) and Table 1.

The results of image prior-based methods suffer from color shift in heavy hazy area or white image patches, and five deep learning-based methods avoid color shift but produce poor-effect dehazing outputs. However, the results of our FWB-Net are haze-free as well as avoiding color shift. It shows our method has the most outstanding performance of all the compared methods.

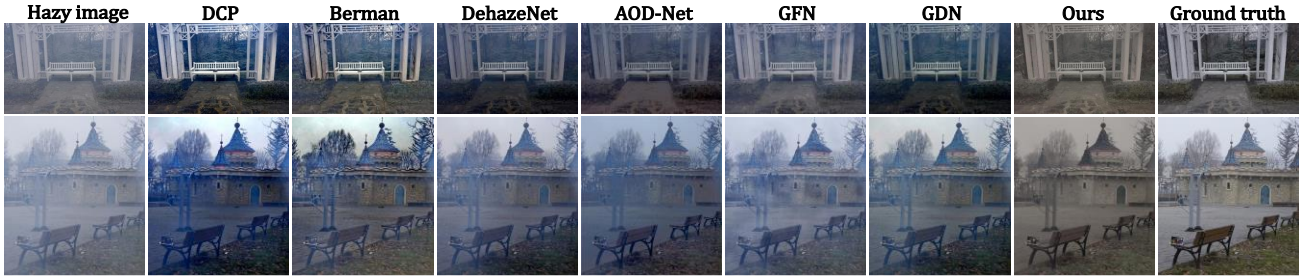

**Fig. 4.** Evaluation on real-world hazy dataset O-Haze. Results demonstrate that our method can get clean dehazing results with restraining color shift. While others get unclean dehazing results, or suffer from color shift.

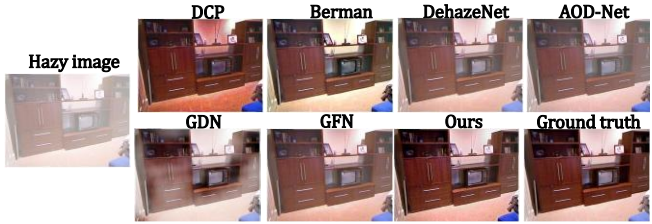

(a)

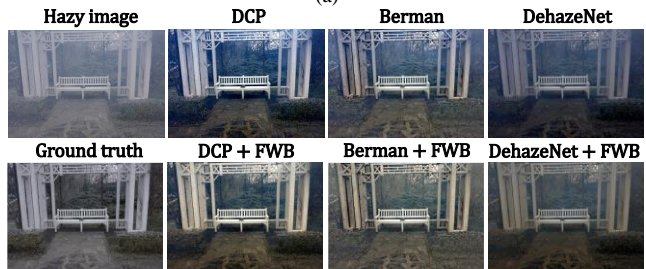

(b)

**Fig. 5. (a)** Evaluation on synthetic hazy dataset NYU-Depth with $\beta = 0.35$. **(b)** Applying FWB-Module to classical image dehazing methods. Results demonstrate that FWB-Module can correct the color shift of outputs of other ASM-based dehazing methods.

**Table 1.** Quantitative comparisons for different methods. Our method ranks all first of those 3 indexes on the 2 datasets

| Model | NYU-Depth | | | O-Haze | | |
|---|---|---|---|---|---|---|
| | P | S | C | P | S | C |
| DCP [4] | 17.3 | 0.785 | 10.8 | 16.6 | 0.735 | 20.8 |
| Berman [5] | 16.6 | 0.763 | 13.1 | 16.6 | 0.750 | 17.1 |
| DehazeNet [6] | 14.7 | 0.749 | 14.2 | 16.2 | 0.666 | 17.3 |
| AOD-Net [8] | 12.8 | 0.688 | 18.5 | 17.1 | 0.664 | 15.8 |
| GFN [9] | 18.6 | 0.758 | 10.3 | 18.1 | 0.671 | 13.4 |
| GDN [10] | **18.9** | 0.763 | 9.9 | 18.7 | 0.672 | 12.6 |
| Ours | **18.9** | **0.794** | **8.7** | **19.7** | **0.763** | **9.4** |

Note: P, S, C, denotes PSNR, SSIM, CIEDE2000 respectively.

**Table 2.** Quantitative results for applying FWB-Module to classical ASM-based image dehazing algorithms. It shows FWB-Module can improve the effect of other dehazing methods.

| Model | Original | | | Add FWB-Module | | |
|---|---|---|---|---|---|---|
| | P | S | C | P | S | C |
| DCP [4] | 16.6 | 0.735 | 20.8 | 17.0 | 0.741 | 16.7 |
| Berman [5] | 16.6 | 0.750 | 17.1 | 16.9 | 0.754 | 16.3 |
| DehazeNet [6] | 16.2 | 0.666 | 17.3 | 16.5 | 0.680 | 15.5 |

Although our work focuses on improving visibility, it also does well in quantitative comparisons. Peak Signal to Noise Ratio (PSNR) [18] and Structural Similarity Index Measure (SSIM) [19] are commonly used in evaluating dehazing results. CIEDE2000 [20] measures accurately the color difference between two images and generates values in the range [0,100], with smaller values indicating better color shift avoiding ability, which can evaluate the level of color restoration. We calculate PSNR, SSIM, CIEDE2000 of all comparing methods and our method and provide the result in Table 1. Our FWB-Net ranks all first in the whole 3 indexes on the synthetic dataset. It implies that our proposed method is effective.

### 3.3. Evaluations on real-world hazy image dataset

We evaluate our method and six comparing methods on real-world hazy image dataset O-Haze [21]. It has 45 images, we choose 40 of them to train our model from scratch and 5 images are used as testing samples. We present some of the results in Fig. 4.

In Fig. 4, we can see that our results not only are the closest to the ground truth, but also performs best in avoiding color shift. After dehazing, the results of the compared methods are blue, while ours restore the color. Moreover, in Table 1, our method performs the best in PSNR, SSIM and CIEDE2000 on O-Haze dataset. Therefore, the qualitative and quantitative results show that our method leads to a superior performance in real world hazy condition.

### 3.4. The correction of FWB-Module to color shift of classical image dehazing methods

Without FWB-Module, the PSNR of FWB-Net will reduce 0.9 and 0.7 on NYU-Depth and O-Haze dataset, respectively. And as an independent color shift adjustment module, FWB-Module can also replace the atmospheric light factor estimation module in other ASM-based image dehazing methods. It can correct serious color shift results of many classic image dehazing methods. We apply FWB-Module to DCP [4], Berman [5], DehazeNet [6] and test them on O-Haze dataset. The results are shown in Fig. 5 (b) and Table 2. It can be found that compared with the classical atmospheric light estimation methods, FWB-Module can significantly reduce the color shift of outputs. Therefore, the comparison results show our FWB-module is effective, robust and portable.

## 4. CONCLUSIONS

In this paper, first, we propose a new non-homogeneous atmospheric scattering model (NH-ASM) to precisely model the complex illumination conditions in real hazy scenes. Then, we design a new front white balance module (FWB-Module) and its exclusively training loss function (FWB-Loss), respectively. FWB-Module can correct the color shift of dehazing result by estimating atmospheric light factor. And FWB-Loss imposes penalty on color shift while training FWB-Module. We further incorporate the proposed FWB-Module and FWB-Loss to form a unified dehazing network called FWB-Net, which can restrain the color shift as well as dehazing. Experimental results show our method can get outstanding clean dehazing outputs without color shift.